\RequirePackage[T1]{fontenc}
\documentclass[journal]{IEEEtai}
\usepackage[utf8]{inputenc}
\usepackage{amsmath,amssymb,bm}
\usepackage{graphicx,booktabs,array,tabularx,multirow}
\usepackage{flafter}
\usepackage[table]{xcolor}
\usepackage{tikz}
\usetikzlibrary{arrows.meta,positioning,fit,calc}
\usepackage[nocompress]{cite}
\usepackage[colorlinks=true,urlcolor=blue!55!black,linkcolor=black,citecolor=black]{hyperref}
\hypersetup{pdfauthor={},pdftitle={},pdfsubject={},pdfkeywords={}}
\newcommand{\method}{KolTa}
\newcommand{\R}{\mathbb R}
\newcommand{\Pset}{\mathcal P}
\newcommand{\Lset}{\mathcal L}
\newcommand{\Uset}{\mathcal U}
\newcommand{\Oset}{\mathcal O}
\newcommand{\Tref}{T_{\mathrm{ref}}}
\newcommand{\ph}{\mathrm{phys}}

\newcommand{\meany}{\widehat y}

\definecolor{bestpink}{HTML}{F7B7D6}
\definecolor{secondpink}{HTML}{FCE4F0}
\newcommand{\best}[1]{\cellcolor{bestpink}\textbf{#1}}
\newcommand{\secondbest}[1]{\cellcolor{secondpink}#1}
\newcolumntype{Y}{>{\raggedright\arraybackslash}X}

\begin{document}
\bstctlcite{IEEEexample:BSTcontrol}
\title{KolTa: A Physics-Informed Active Learning Framework for Formulation Discovery with Limited Data}
\author{Boyong~Tang, Lei~Yang, and Tao~Wang%
\thanks{\textit{Corresponding author: Tao Wang.}}%
\thanks{Boyong Tang and Tao Wang are with the School of
Artificial Intelligence, Sichuan University, Chengdu, China
(e-mail: boyong.tang.math.ai@gmail.com; twangnh@gmail.com).}%
\thanks{Lei Yang is with the School of Aeronautics and
Astronautics, Sichuan University, Chengdu, China
(e-mail: yang-lei@scu.edu.cn).}%
}
\markboth{}{}
\maketitle
\begin{abstract}
Finding high-performing formulations is difficult when few have been measured and experimental budgets are limited. We propose \method{}, an active learning framework that selects batches to discover the top 5\% of formulations at a target temperature under a fixed query budget. \method{} combines a three-stage predictor of component contributions, role-conditioned interactions, and anchored temperature response with task-specific soft physics constraints. Its Physics-Residual-Calibrated Upper Confidence Bound (PRC-UCB) acquisition rule combines predicted utility, ensemble disagreement, and a target-temperature physics residual, with nonnegative exploration weights fitted to out-of-fold signed errors. We evaluate the framework on three measured tasks and one simulated task covering binary-mixture viscosity, drug solubility, ionic-liquid viscosity, and electrolyte conductivity. \method{} ranks first or second in mean area under the learning curve (AULC) on every task and outperforms seven of the eight comparator strategies in paired tests, winning 75 of 78 paired comparisons across tasks and seeds against those seven strategies. Across the three measured tasks, its largest gap to the best mean AULC is 0.032, compared with at least 0.056 for every comparator. On Drug, reaching 80\% average recovery of the remaining targets across drugs requires an estimated median of 348 additional formulation queries, about 18\% fewer than the fastest comparator. Researchers can adapt the framework to diverse formulation problems through component roles and physical constraints.
\end{abstract}
\begin{IEEEImpStatement}
Developing multi-component formulations such as drug--solvent systems, ionic-liquid mixtures, and battery electrolytes requires screening large candidate spaces under tight experimental budgets. \method{} helps formulation teams prioritize which candidates to prepare and measure next by combining multi-temperature records with component roles and governing temperature relations. Across four campaigns spanning viscosity, solubility, and conductivity, the framework consistently ranks in the top two discovery strategies without retuning its acquisition rule for each domain, cutting median additional queries to recover 80\% of the remaining top-5\% drug--solvent systems by nearly one-fifth relative to the fastest baseline. Practitioners can adapt its role-conditioned interactions and soft physical constraints to accelerate early-stage laboratory screening.
\end{IEEEImpStatement}
\begin{IEEEkeywords}
Active learning, Bayesian optimization, formulation discovery, mixture modeling, physics-informed learning.
\end{IEEEkeywords}
\section{Introduction}\label{sec:intro}
\IEEEPARstart{M}{ulti-component} formulations combine several chemical species in tunable proportions, as when dissolving ibuprofen in an ethanol--water cosolvent mixture, blending ionic liquids for low viscosity, or formulating a $\mathrm{LiPF_6}$--carbonate battery electrolyte~\cite{diffmix,dong2007ilthermo,bao2024solubility}. Because properties depend jointly on molecular structures, roles, compositions, and temperature, varying the species and ratios creates vast candidate spaces that are costly to prepare and measure~\cite{chew2025mixtures}. Predictive surrogates help prioritize this search~\cite{lookman2019active}, yet their practical value in early-stage campaigns depends on how reliably they rank untested formulations under a strict budget.

Existing mixture models include approaches tailored to specific chemical systems~\cite{diffmix,yang2026bamboo,sanchezmedina2023gibbs} and general architectures adaptable across prediction tasks~\cite{deepsets,molsets,solvgnn}, alongside recent benchmarks and pretrained representations~\cite{chemixhub,bai2026chemworld}. In early-stage discovery, however, only a small initial set of formulations has been measured, and predictions can deteriorate for candidates whose property values lie outside the training range~\cite{segal2025known}.

Active learning (AL) and Bayesian optimization (BO) address this challenge by iteratively selecting informative experiments and updating the surrogate~\cite{lookman2019active,cameo,noh2024robotic,ma2025electrolyte}. For temperature-dependent properties, closed-loop discovery faces two coupled challenges. First, once a formulation is prepared (such as an ibuprofen--cosolvent mixture), measuring it across several temperatures (e.g., 293.15--313.15~K) is routine, whereas engineering design ranks candidates at a fixed operating temperature ($T^*=303.15$~K). Temperature can be concatenated with a mixture representation, as in the benchmark variants studied by Sanchez Medina et al. and CheMixHub~\cite{sanchezmedina2023gibbs,chemixhub}, without explicitly imposing a thermodynamic constraint on temperature derivatives. Parameterizing outputs directly through empirical equations instead restricts the temperature response to a chosen functional form~\cite{diffmix,chemixhub,bai2026chemworld}. Second, greedy selection ranks candidates by predicted utility, whereas UCB and expected improvement (EI) also use model-predicted uncertainty~\cite{srinivas,jones}. Uncertainty miscalibration can affect Bayesian optimization~\cite{deshpande}, motivating acquisition rules that account for observed prediction errors.

We propose \method{}, a physics-informed active learning framework that couples a three-stage predictor with error-calibrated acquisition. The predictor decomposes properties into component base contributions, role-conditioned interactions, and an anchored temperature response guided by a soft physics derivative constraint. Its Physics-Residual-Calibrated Upper Confidence Bound (PRC-UCB) rule combines predicted utility with ensemble disagreement and the target-temperature physics residual, which measures local departure from the governing temperature relation. Nonnegative margin weights are fitted each round to out-of-fold (OOF) signed errors by minimizing an asymmetric pinball loss, calibrating how strongly each signal contributes to the next batch.

We evaluate discovery by the fraction of top-5\% formulations recovered as queries accumulate, measured by the area under the learning curve (AULC) and the queries to reach 30\% and 80\% recovery ($q_{30}$ and $q_{80}$), and require both high per-task discovery efficiency and cross-task consistency across distinct chemical domains. Our contributions are as follows.

\begin{itemize}
\item We design a three-stage mixture predictor that separates reference-temperature base and interaction readouts from an anchored temperature network guided by soft derivative constraints. Under common greedy selection, this predictor ranks first in mean AULC and $q_{80}$ on both ablation tasks.
\item We introduce PRC-UCB, which fits exploration weights for ensemble disagreement and the target-temperature physics residual to OOF signed errors. Calibration assigns positive weight to the residual in 115 of 120 rounds across four tasks and improves mean AULC over disagreement-only calibration on both ablation tasks (five of six runs).
\item Across three measured tasks and one simulated task spanning binary mixtures, drug solutions, ionic liquids, and electrolytes, \method{} ranks first or second in mean AULC on every task, wins 75 of 78 paired comparisons across tasks and seeds against seven of the eight comparator strategies, cuts median queries to 80\% recovery on Drug by 18\%, and has the smallest worst-case AULC gap across the measured tasks (0.032 vs.\ $\ge$0.056).
\end{itemize}

\section{Background and Related Work}\label{sec:related}
\textbf{Mixture representation and property prediction.} Mixture models range from domain-specific workflows to general graph architectures. Chew et al.~\cite{chew2025mixtures} use molecular-dynamics-generated mixture properties to benchmark descriptor, Set2Set, and graph models alongside active learning, while GeoMix models molecular geometry and interactions for electrolyte conductivity~\cite{geomix2025}. For general mixtures, Deep Sets aggregates component embeddings~\cite{deepsets}, MolSets combines molecular graphs with learned aggregation~\cite{molsets}, and SolvGNN passes messages between components~\cite{solvgnn}. CheMixHub benchmarks descriptor and graph models across mixture datasets, including Arrhenius-parameter predictors~\cite{chemixhub}. Compared with a direct temperature-concatenation readout, \method{} separates role-conditioned reference-temperature readouts from an anchored temperature response.

\textbf{Physics-informed learning.} Physics-informed neural networks add a governing-equation residual to the training loss~\cite{raissi}. Chem World's Mixture-PINN imposes soft constraints on component contributions, symmetry, and temperature monotonicity~\cite{bai2026chemworld}. Physical relations can also parameterize outputs directly, as in DiffMix (Vogel--Fulcher--Tammann and Redlich--Kister relations)~\cite{diffmix} and BAMBOO-Mixer~\cite{yang2026bamboo}. Rather than restricting predictions to a closed-form curve, \method{} anchors its temperature subnetwork at $\Tref$ and regularizes its derivative via a task-specific residual, preserving flexible reference-temperature predictions while supplying a local residual for acquisition.

\textbf{Active learning and Bayesian optimization for discovery.} Closed-loop discovery combines surrogate models with sequential selection across materials and electrolyte campaigns~\cite{lookman2019active,cameo,noh2024robotic,ma2025electrolyte}. Common acquisition rules include UCB~\cite{srinivas} and expected improvement (EI)~\cite{jones,ament2023unexpected}. Deep ensembles provide one approach to estimating predictive uncertainty~\cite{ensemble}. Recent work recalibrates predictive distributions via cross-validation~\cite{deshpande}, applies conformal prediction to PDE residuals~\cite{gopakumar}, weights acquisition by prior beliefs or stoichiometric priors~\cite{hvarfner2022pibo,wakabayashi2026pibo}, or penalizes energy-balance violations~\cite{dong2026physicsbo}. Existing mixture AL workflows screen simulation-derived targets with standard EI or greedy rules~\cite{chew2025mixtures}, fixed-temperature solubility with GP-EI~\cite{noh2024robotic}, or battery cycling with multi-kernel GP model averaging and EI~\cite{ma2025electrolyte}, whereas \method{} exploits multi-temperature formulation records and calibrates nonnegative weights for ensemble disagreement and the target-temperature physics residual to OOF signed errors.

\section{Method}
Starting from a few labeled formulations, \method{} couples a three-stage predictor, soft physics training, and PRC-UCB acquisition to find top candidates at a target temperature.

\begin{figure}[!htbp]
\centering
\includegraphics[width=\columnwidth]{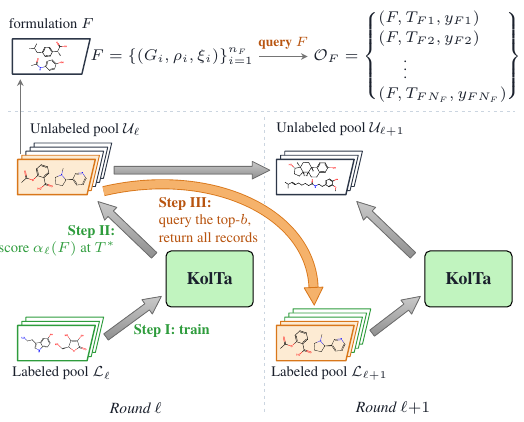}
\caption{Closed-loop active learning with \method{}. Each card is a formulation $F$ with component graphs, roles, and compositions (such as ibuprofen in a binary solvent mixture); querying $F$ returns its records $\Oset_F$ in \eqref{eq:records} (top). Each round trains on $\Lset_\ell$ (Step~I), scores $F\in\Uset_\ell$ by $\alpha_\ell(F)$ at $T^*$ via \eqref{eq:acq} (Step~II), and queries the top-$b$ batch to form $\Lset_{\ell+1}$ and $\Uset_{\ell+1}$, restricted to the scheduled drug group on Drug (Step~III).}
\label{fig:loop}
\end{figure}

\begin{figure*}[t]
\centering
\includegraphics[width=\textwidth]{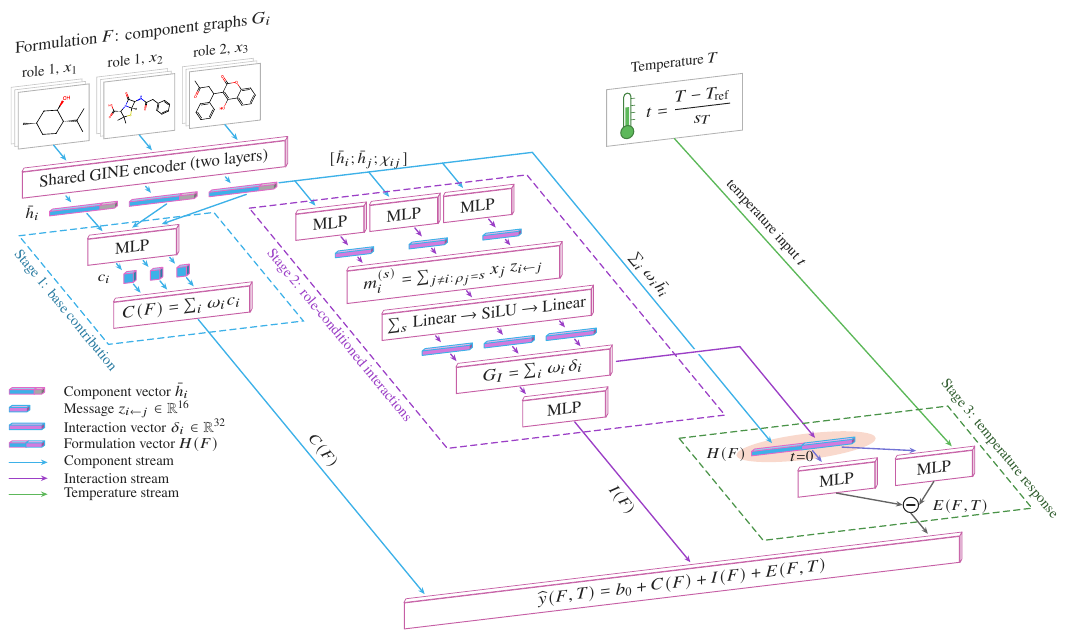}
\caption{Three-stage predictor of \method{}, illustrated with three components in two roles. A shared GINE encoder maps each graph to $h_i\in\R^{32}$ (forming $\bar h_i=[h_i;d_i]$). Stage~1 applies the MLP $g_C$ to compute $C(F)$ in \eqref{eq:base}. Stage~2 applies role-pair message MLPs $f_{\rho_i\leftarrow\rho_j}$, sums messages by sender role, updates receivers via $\sum_s W_{\rho_i\leftarrow s}\to\operatorname{SiLU}\to O_{\rho_i}$, and maps $G_I(F)$ through the MLP $g_I$ to $I(F)$ in \eqref{eq:msg}--\eqref{eq:inter}. Stage~3 evaluates the MLP $q_\phi$ on $H(F)$ in \eqref{eq:repr} at $t$ and $0$ to obtain $E(F,T)$ with $E(F,\Tref)=0$ in \eqref{eq:temp_effect}--\eqref{eq:pred}; $H(F)$ also feeds the linear physics head $\theta(F)$ for $r(F,T)$ in \eqref{eq:residual}.}
\label{fig:predictor}
\end{figure*}

\subsection{Problem Setting}\label{sec:setting}
\textbf{Formulations and measurement records.} Each formulation $F$ in a finite candidate pool $\Pset$ contains $n_F$ components. Component $i$ has a molecular graph $G_i$, a role $\rho_i$ (e.g., solute, solvent, or salt), and a composition $\xi_i$ (e.g., ibuprofen in an ethanol--water cosolvent mixture; Fig.~\ref{fig:loop}). A property record $(F,T,y)$ gives its value $y$ at temperature $T$, from measurement or simulation. Structures, roles, compositions, and recorded temperatures are known for all candidates; queries return their property values.

\textbf{Queries.} Viscosity~\cite{riccardi2022thermoml,dong2007ilthermo} and solubility~\cite{bao2024solubility} datasets record formulations at different temperatures. We count queries in formulations: querying $F$ returns all its available records,
\begin{equation}\label{eq:records}
 \Oset_F=\{(F,T_{Fn},y_{Fn})\}_{n=1}^{N_F},
\end{equation}
where $N_F$ is the record count in \eqref{eq:records}, and $T_{Fn}$ and $y_{Fn}$ are the temperature and property value of record $n$. All enter training, and $F$ is not queried again.

\textbf{Discovery objective.} Training uses records at different temperatures, and candidates are compared at $T^*$ by their utility $s\,y(F,T^*)$, where $s=+1$ for maximization and $s=-1$ for minimization. The targets are the top 5\% by utility, including ties at the $k$-th ranked value, where $k=\lceil0.05|\Pset|\rceil$ and $|\Pset|$ is the number of candidates. Every candidate has a record at $T^*$ for evaluation; Drug defines targets within each drug (Section~\ref{sec:data}). We aim to find as many targets as possible, as early as possible, within a fixed query budget.

\textbf{Active learning loop.} Starting from $\Lset_0$, each round $\ell$ trains the model on all records of $\Lset_\ell$ and evaluates each unlabeled formulation $F\in\Uset_\ell=\Pset\setminus\Lset_\ell$ by its acquisition score $\alpha_\ell(F)$ (Section~\ref{sec:acq}). We query a batch $\mathcal Q_\ell$ with a fixed size $b$ and update $\Lset_{\ell+1}=\Lset_\ell\cup\mathcal Q_\ell$. This repeats until the budget is reached (Fig.~\ref{fig:loop}).

\subsection{Three-Stage Predictor}\label{sec:predictor}
The predictor first encodes the components, then learns their base contributions, interactions, and temperature response (Fig.~\ref{fig:predictor}). Supplementary Sections~S1--S2 provide implementation and training details.

\textbf{Component encoding.} A shared two-layer GINE network~\cite{hu2020} encodes each component's molecular graph, using atom and bond features, into $h_i\in\R^{32}$. Optional component-level physical properties $d_i$, chosen by task and role, are standardized and concatenated to form $\bar h_i=[h_i;d_i]$ ($\bar h_i=h_i$ when omitted). Drug uses the drug's melting point and each solvent's melting point and reference relative permittivity; NIST, ILThermo, and DiffMix append no additional component properties.

\textbf{Stage 1: base contribution.} The task determines the component set $\mathcal D_F$ for the base readout: all components for bulk viscosity and conductivity, and only the drug solute for Drug (with solvents acting in Stage~2). For each $i\in\mathcal D_F$, a shared MLP $g_C$ maps $\bar h_i$ and role encoding $e(\rho_i)$ (omitted in single-role tasks) to $c_i=g_C([\bar h_i;e(\rho_i)])$, weighted by $\omega_i$:
\begin{equation}\label{eq:base}
 C(F)=\sum_{i\in\mathcal D_F}\omega_i c_i .
\end{equation}
The weights $\omega_i$ in \eqref{eq:base} are given for each task below; the sum is permutation-invariant~\cite{deepsets}.

\textbf{Stage 2: role-conditioned interactions.} We model component interactions via one round of directed message passing~\cite{gilmer2017mpnn,solvgnn}. From sender $j$ to receiver $i$, roles $(\rho_j,\rho_i)$ select the interaction MLP $f_{\rho_i\leftarrow\rho_j}$:
\begin{equation}\label{eq:msg}
 z_{i\leftarrow j}=f_{\rho_i\leftarrow\rho_j}\bigl([\bar h_i;\bar h_j;\chi_{ij}]\bigr)\in\R^{16}.
\end{equation}
Here, $\chi_{ij}$ in \eqref{eq:msg} contains composition inputs: $\chi_{ij}=[x_i;x_j]$ when receiver and sender fractions share a common basis, and $\chi_{ij}=x_j$ otherwise (on Drug, the solvent's mole fraction within the solvent mixture).

Messages of the same interaction type share parameters~\cite{schlichtkrull2018rgcn}. For each receiver $i$, incoming messages are grouped by sender role $s$ and weighted by $x_j$:
\begin{equation}\label{eq:msg_sum}
 m_i^{(s)}=\sum_{j\neq i:\,\rho_j=s}x_j z_{i\leftarrow j}\in\R^{16}.
\end{equation}

Zero-fraction senders contribute nothing to \eqref{eq:msg_sum}. Transforming each role sum by $W_{\rho_i\leftarrow s}$ ($16\to32$), applying SiLU, and projecting via $O_{\rho_i}$ (both bias-free, so $\delta_i=0$ without incoming messages) gives:
\begin{equation}\label{eq:delta}
 \delta_i=O_{\rho_i}\operatorname{SiLU}\Bigl(\sum_s W_{\rho_i\leftarrow s}m_i^{(s)}\Bigr)\in\R^{32}.
\end{equation}

Pooling $\delta_i$ over $\mathcal D_F$ with weights $\omega_i$ yields $G_I(F)$, which a bias-free SiLU MLP $g_I$ maps to $I(F)$:
\begin{equation}\label{eq:inter}
 G_I(F)=\sum_{i\in\mathcal D_F}\omega_i\delta_i,\qquad I(F)=g_I\bigl(G_I(F)\bigr),
\end{equation}
so $G_I(F)=0$ and $I(F)=0$ when no interactions are active in \eqref{eq:inter}. Concatenating the pooled Stage~1 features with $G_I(F)$ forms
\begin{equation}\label{eq:repr}
 H(F)=\Bigl[\textstyle\sum_{i\in\mathcal D_F}\omega_i\bar h_i\,;\ G_I(F)\Bigr],
\end{equation}
which supplies Stage~3 and the physics head (Section~\ref{sec:physics}).

\textbf{Stage 3: anchored temperature response.} We use $t=(T-\Tref)/s_T$, with reference temperature $\Tref$ and positive temperature scaling factor $s_T$ fixed for each task. A network $q_\phi$ takes $H(F)$ and $t$. Subtracting its output at $t=0$ expresses the temperature effect relative to $\Tref$:
\begin{align}
 E(F,T)&=q_\phi\bigl(H(F),t\bigr)-q_\phi\bigl(H(F),0\bigr),\label{eq:temp_effect}\\
 \meany(F,T)&=b_0+C(F)+I(F)+E(F,T),\label{eq:pred}
\end{align}
where $b_0$ is a trainable bias. Since $E(F,\Tref)=0$ in \eqref{eq:temp_effect}, $b_0+C(F)+I(F)$ in \eqref{eq:pred} predicts the property at $\Tref$, while $E(F,T)$ predicts its change with temperature. Candidates are scored at $t^*=(T^*-\Tref)/s_T$.

\textbf{Task instances.} Viscosity uses one component role and message function, $\chi_{ij}=[x_i;x_j]$, and mole-fraction weights $\omega_i=x_i$ for all components. The architecture thus handles both binary and ternary mixtures. Drug uses only solvent-to-solute messages, with $\chi_{ij}=x_j$ defined within the solvent mixture; the solute's saturated mole fraction is the prediction target. Only the solute enters the readout, with $\omega=1$. Electrolytes use three message types (salt$\leftarrow$solvent, solvent$\leftarrow$salt, and solvent$\leftarrow$solvent), each with its own $f$ and $W$. Here, $\chi_{ij}=[x_i;x_j]$ and readout weights use overall mole fractions.

\subsection{Soft Physics Constraint and Training}\label{sec:physics}
A physics-informed neural network (PINN) approach adds a derivative-based physics loss to guide how the predicted property changes with temperature~\cite{raissi}.

\textbf{Physics residual.} For task $k$, $g_k(T;\theta)$ is the temperature derivative specified by the physical relation. A linear output layer, the physics coefficient head, predicts $\theta(F)$ from $H(F)$. Fixed functions keep bounded coefficients within their allowed ranges. The residual compares the network derivative with this relation:
\begin{equation}\label{eq:residual}
 r(F,T)=\partial_t\meany(F,T)-s_T\,g_k\bigl(T;\theta(F)\bigr).
\end{equation}
Automatic differentiation gives $\partial_t\meany$ in \eqref{eq:residual}; the factor $s_T$ converts $g_k$ to the same coordinate, since $\partial_t\meany=s_T\partial_T\meany$.

\textbf{Relations used in this study.} For log viscosity, the Arrhenius form~\cite{chemixhub} gives $g_k=-\theta(F)/T^2$; for log solubility, the van't Hoff form~\cite{moreno2022solubility} gives $g_k=\theta(F)/T^2$. The learned scalar $\theta(F)$ can have either sign. For log electrolyte conductivity, the Vogel--Fulcher--Tammann relation~\cite{diffmix} gives $g_k=\beta(F)/(T-T_0(F))^2$, with $\beta(F)\ge0$ and $0<T_0(F)<233.15$~K. Each formulation's predicted coefficients are constant across temperatures.

\textbf{Joint training.} The data loss uses all measured records, while the physics loss uses five fixed temperatures $\{T_c\}_{c=1}^{5}$ per task (Table~\ref{tab:data}) without requiring measurements there. For the current labeled set $\Lset$,
\begin{align}
 \mathcal L_{\mathrm{data}}
 &=\frac{1}{|\Lset|}\sum_{F\in\Lset}\frac{1}{N_F}
   \sum_{n=1}^{N_F}\bigl(\meany(F,T_{Fn})-y_{Fn}\bigr)^2,\label{eq:loss_data}\\
 \mathcal L_{\ph}
 &=\frac{1}{|\Lset|}\sum_{F\in\Lset}\frac{1}{5}
   \sum_{c=1}^{5}r(F,T_c)^2,\label{eq:loss_phys}\\
 \mathcal L&=\mathcal L_{\mathrm{data}}+\lambda\mathcal L_{\ph},\label{eq:loss_total}
\end{align}
where $\lambda>0$ weights $\mathcal L_{\ph}$ in \eqref{eq:loss_total}. Averaging errors within each formulation in \eqref{eq:loss_data}--\eqref{eq:loss_phys} prevents those with more records from receiving greater weight.

Anchoring $E(F,\Tref)=0$ and excluding $b_0$, $C(F)$, and $I(F)$ from $r(F,T)$ means their readout layers are updated directly by $\mathcal L_{\mathrm{data}}$, while $\mathcal L_{\ph}$ regularizes the shared encoder and interaction modules through $H(F)$. Moreover, because the physics head $\theta(F)$ enters the objective solely through $r(F,T)$ in \eqref{eq:loss_phys}, $\mathcal L_{\ph}$ is required to train $\theta(F)$ so that the target-temperature residual $r(F,T^*)$ is well-defined for acquisition; when $T^*=\Tref$ ($t^*=0$, measured tasks), $E(F,T^*)=0$, whereas on DiffMix ($t^*=1$), $E(F,T^*)$ also predicts the shift from $\Tref$ to $T^*$.

The predictor's parameters are trained jointly with Adam. The learning rate, weight decay, and $\lambda$ are selected by target-temperature mean absolute error (MAE) on an inner validation split grouped by formulation. Five ensemble members are then trained on bootstrap samples of formulations, keeping each formulation's records together and counting its loss once per draw.

\subsection{PRC-UCB Acquisition}\label{sec:acq}
PRC-UCB adds a calibrated exploration margin to predicted utility at $T^*$. The margin combines ensemble disagreement and the physics residual, with weights refitted each round using labeled formulations.

\textbf{Prediction and exploration signals.} Let $\meany_m(F,T^*)$ and $r_m(F,T^*)$ be the prediction and residual of member $m$, with $M=5$. Their prediction mean gives $\mu(F)$, and their sample standard deviation gives $u(F)$~\cite{ensemble}. The physics signal averages absolute residuals from \eqref{eq:residual}, avoiding cancellation between opposite signs:
\begin{equation}\label{eq:phys_signal}
 v(F)=\frac{1}{M}\sum_{m=1}^{M}\left|r_m(F,T^*)\right|.
\end{equation}

\textbf{Out-of-fold errors.} We split labeled formulations into five folds, keeping each formulation's records together. Hyperparameter selection and ensemble training on the other four folds give out-of-fold (OOF) predictions $\tilde\mu(F)$ and signals $\tilde u(F)$ and $\tilde v(F)$ from \eqref{eq:phys_signal} at $T^*$. The held-out labels then give the signed utility error
\begin{equation}\label{eq:oof_err}
 e(F)=s\bigl(y(F,T^*)-\tilde\mu(F)\bigr),
\end{equation}
where $s$ in \eqref{eq:oof_err} is defined in Section~\ref{sec:setting}. Positive errors indicate underestimated utility for either objective direction.

\textbf{Calibration of the exploration margin.} We fit the coefficients of the linear exploration margin $c_u u(F)+c_v v(F)$ to the out-of-fold signed errors by minimizing an asymmetric pinball loss~\cite{koenker}:
\begin{equation}\label{eq:pinball}
 (c_u,c_v)=\operatorname*{arg\,min}_{c_u,c_v\ge0}\ \sum_{F\in\Lset}\rho_\tau\bigl(e(F)-c_u\tilde u(F)-c_v\tilde v(F)\bigr),
\end{equation}
where $\rho_\tau(z)=\max(\tau z,(\tau-1)z)$ in \eqref{eq:pinball}. We set $\tau=0.8$, so a margin that falls short of the signed error receives four times the per-unit penalty of a margin that exceeds it. Either coefficient can be zero; $c_v=0$ leaves only ensemble disagreement.

To handle multiple solutions, we first scale each out-of-fold signal by its root mean square over $\Lset$. For $a\in\{u,v\}$, define
\begin{equation}\label{eq:rms_scale}
 \sigma_a=\sqrt{\frac{1}{|\Lset|}\sum_{F\in\Lset}\tilde a(F)^2},\qquad
 a'(F)=\frac{\tilde a(F)}{\sigma_a}.
\end{equation}
A signal with $\sigma_a=0$ in \eqref{eq:rms_scale} is omitted and receives a zero coefficient. We then minimize pinball loss using $u'$ and $v'$. If several coefficient pairs minimize the loss, we select the pair $(c'_u,c'_v)$ with the smallest Euclidean norm, $\sqrt{(c'_u)^2+(c'_v)^2}$. Finally, we convert the coefficients back:
\begin{equation}\label{eq:unscale}
 c_u=\frac{c'_u}{\sigma_u},\qquad c_v=\frac{c'_v}{\sigma_v},
\end{equation}
for the retained signals in \eqref{eq:unscale}. This preserves the fitted margin, $c'_u u'+c'_v v'=c_u\tilde u+c_v\tilde v$, and makes the choice among minimizing solutions independent of signal units.

\textbf{Candidate selection.} With the calibrated coefficients, the acquisition score is
\begin{equation}\label{eq:acq}
 \alpha(F)=s\,\mu(F)+c_u u(F)+c_v v(F),
\end{equation}
and candidates are ranked by \eqref{eq:acq} to select the next batch of $b$ formulations. Drug applies this ranking within each scheduled drug group (Section~\ref{sec:data}).

\section{Experimental Design}\label{sec:experiments}
We evaluate \method{}'s formulation discovery performance under fixed query budgets, compare predictors using a common greedy rule, and examine the progression from greedy selection to disagreement-only and physics-residual-calibrated acquisition.

\subsection{Tasks and Data}\label{sec:data}
We evaluate the framework through retrospective discovery on three measured datasets and one simulated dataset (Table~\ref{tab:data}). The measured records are distributed by CheMixHub~\cite{chemixhub}: binary-mixture viscosity from NIST/TRC SOURCE~\cite{kazakov2012source} (NIST), drug-solubility records in one or two solvents compiled by Bao et al.~\cite{bao2024solubility}, from which we retain 86 drug groups (Drug), and ionic-liquid mixture viscosity from ILThermo~\cite{dong2007ilthermo}. DiffMix~\cite{diffmix} provides simulated conductivities of non-aqueous lithium electrolytes. This task evaluates the framework with a different temperature relation and three interaction types.

\begin{table*}[t]
\centering
\caption{Tasks and query budgets. The three measured datasets are distributed by CheMixHub~\cite{chemixhub}. The budget comprises the initial labeled set and $R$ rounds of $b$ formulation queries. Each task uses five collocation temperatures within the stated range. Temperatures enter the predictor as $t=(T-\Tref)/s_T$, with $\Tref=T^*$ ($t^*=0$) and $s_T=10$~K for viscosity and 50~K for solubility; DiffMix uses the center of its simulated range, $\Tref=268.15$~K, and $s_T=25$~K ($t^*=1$). VFT: Vogel--Fulcher--Tammann.}
\label{tab:data}
\footnotesize
\begin{tabular}{@{}llrrcll@{}}
\toprule
Task & Property (goal) & Pool / locked & Records & $T^*$ (K) & Budget & Relation, collocation (K)\\
\midrule
NIST~\cite{kazakov2012source} & Binary-mixture viscosity (max.) & 15,575 / 3,894 & 50,429 & 298.15 & $200+10\times80$ & Arrhenius, 293.15--313.15\\
Drug~\cite{bao2024solubility} & Drug solubility, per drug (max.) & 2,159 / 524 & 23,688 & 303.15 & $300+10\times80$ & van't Hoff, 293.15--313.15\\
ILThermo~\cite{dong2007ilthermo} & Ionic-liquid mixture viscosity (min.) & 1,940 / 486 & 17,072 & 313.15 & $150+10\times30$ & Arrhenius, 303.15--323.15\\
DiffMix~\cite{diffmix} (sim.) & Electrolyte conductivity (max.) & 1,471 / 368 & 11,022 & 293.15 & $100+10\times40$ & VFT, 243.15--293.15\\
\bottomrule
\end{tabular}
\end{table*}

Formulations are assigned to a discovery pool $\Pset$ and a locked set without consulting property values. NIST, ILThermo, and DiffMix use an 80/20 split. Drug retains formulation assignments after drug-group filtering, resulting in 2,159 discovery and 524 locked formulations (Supplementary Section~S2). The locked set serves only to evaluate final prediction error (Supplementary Section~S4). All records of a formulation stay together. All predictive methods use the source simplified molecular-input line-entry system (SMILES) strings. For Drug, they also use the drug's melting point and each solvent's melting point and reference relative permittivity. NIST, ILThermo, and DiffMix use no such additional component properties. Targets $y$ are base-10 logarithms of viscosity in cP, saturated solute mole fraction, or conductivity in S/cm.

\textbf{Initialization and queries.} All methods share a random initial set $\Lset_0\subset\Pset$, drawn without consulting property values, and $R$ rounds of $b$ queries, giving $B=bR$ new queries. Each evaluated method--task pair uses seeds 0, 1, and 2 to determine initialization, training, and out-of-fold splits.

\textbf{Drug task.} Targets are defined within each drug, and initialization covers all drugs. Queries follow a fixed, balanced drug schedule, selecting the highest-scoring available formulation within each scheduled drug.

\subsection{Comparators and Ablations}\label{sec:comparators}
We compare \method{} with MolSets~\cite{molsets}, SolvGNN~\cite{solvgnn} (hidden width 128), a Gaussian process (GP)~\cite{rasmussen2006gaussian}, and XGBoost~\cite{chen2016xgboost}. GP and XGBoost use composition-weighted RDKit descriptors and temperature. GP-EI uses an automatic-relevance-determination squared-exponential kernel and sequential batch construction with a Kriging believer~\cite{ginsbourger2010kriging} and log expected improvement~\cite{jones,ament2023unexpected}.

Neural and tree comparators use five bootstrap members and inner validation. Their UCB weight $\gamma$ is set using formulation-grouped five-fold OOF predictions each round. Neural comparators share the learning-rate grid $\{3\times10^{-4},10^{-3}\}$ and weight-decay grid $\{0,10^{-4}\}$ of \method{}, which also selects $\lambda\in\{0.01,0.1,1\}$. Their UCB score is $s\,\mu(F)+\gamma\,u(F)$, where $\gamma$ is the ratio of the root mean squares of OOF errors and $u$, clipped to $[0.25,4]$.

The three measured tasks evaluate all nine strategies, pairing MolSets, SolvGNN, and XGBoost with UCB and greedy selection ($s\,\mu(F)$), and GP with EI and greedy selection. DiffMix evaluates six strategies: \method{}, each comparator family's primary uncertainty rule (MolSets-UCB, SolvGNN-UCB, GP-EI, and XGBoost-UCB), and XGBoost-Greedy.

On Drug and ILThermo, A1 refits the exploration margin in \eqref{eq:pinball} using ensemble disagreement $u$ alone ($c_v=0$), while A2 uses greedy selection ($c_u=c_v=0$). Because the physics head $\theta(F)$ is trained solely through $\mathcal L_{\ph}$ in \eqref{eq:loss_phys}, all \method{} variants retain physics-informed training along their own trajectories: comparing A2 with the greedy comparators isolates the predictor under a common rule, whereas comparing A2, A1, and \method{} tracks the progression from greedy to calibrated $u$-only and residual-augmented acquisition.

\subsection{Evaluation and Statistics}\label{sec:eval}
\textbf{Discovery metric.} Let $\mathcal H$ be the fixed top-5\% target set (Section~\ref{sec:setting}) and $\mathcal Q(q)$ the formulations selected in $q$ new queries. Active recall measures recovery of targets outside the initial set:
\begin{equation}\label{eq:recall}
 A(q)=\frac{|\mathcal Q(q)\cap(\mathcal H\setminus\Lset_0)|}
 {|\mathcal H\setminus\Lset_0|}.
\end{equation}
For Drug, we use macro active recall: at each acquisition round, we evaluate \eqref{eq:recall} within each drug and take an equal-weight average over groups with targets remaining after initialization.

The primary measure, full-budget AULC, integrates $A(q)$ in \eqref{eq:recall} over new queries and normalizes by $B$:
\begin{equation}\label{eq:aulc}
 \mathrm{AULC}=\frac{1}{B}\sum_{r=1}^{R}(q_r-q_{r-1})
 \frac{A(q_r)+A(q_{r-1})}{2},
\end{equation}
where $q_0=0$, $q_R=B$, and $A(0)=0$ in \eqref{eq:aulc}. Higher AULC means more targets are found earlier. The Random baseline uses three replays per task. It selects unqueried formulations uniformly at random; on Drug, selection is within the drug group assigned by the balanced schedule.

\textbf{Secondary measures.} $q_{30}$ and $q_{80}$ estimate the new formulation queries needed to reach 30\% and 80\% active recall by linear interpolation between batches. We report the median among runs that reach each level and their count; lower values indicate faster discovery.

\textbf{Statistics.} We use paired comparisons of full-budget AULC, matching runs by task and seed: twelve pairs for strategies run on all four tasks, nine for the other greedy strategies. We report support over these evaluated runs when one-sided exact sign and Wilcoxon signed-rank tests~\cite{wilcoxon1945}, each Holm-adjusted~\cite{holm1979} across eight comparator strategies, both yield $p<0.05$. Medians and ranges summarize paired differences. We summarize the observed differences in query counts, task rankings, and ablation results.

\textbf{Cross-task consistency.} A general formulation discovery framework must perform reliably across diverse chemical domains without task-specific algorithm selection. We therefore evaluate cross-task consistency across NIST, Drug, and ILThermo (where all nine strategies are run) using two complementary criteria: per-run top-three rank frequency across the nine task--seed runs and the worst-case mean-AULC gap. For each task, we subtract each strategy's three-run mean AULC from the highest mean among the nine strategies, and take the largest gap across the three tasks; a smaller value indicates a smaller worst-case shortfall. We repeat this calculation across Drug and ILThermo for PRC-UCB, $u$-only A1, and greedy A2, referenced to the highest mean AULC among these three variants on each task.

\section{Results}\label{sec:results}

\subsection{Discovery under a Fixed Budget}\label{sec:discovery}
\begin{figure}[!htbp]
\centering
\includegraphics[width=\columnwidth]{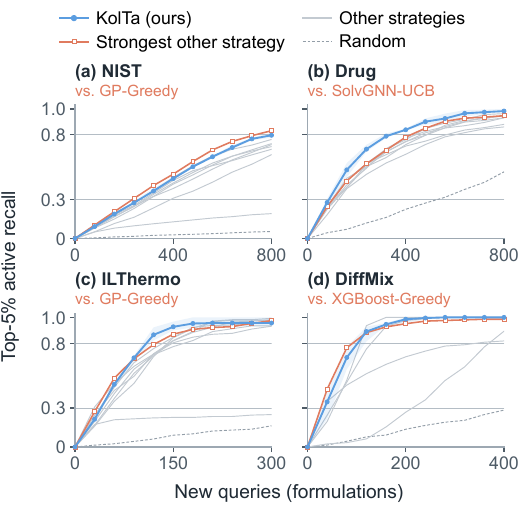}
\caption{\method{} ranks among the top two strategies on all four tasks, although the strongest comparator varies by task. Three-run mean active recall (shaded range for \method{}) on (a) NIST, (b) Drug, (c) ILThermo, and (d) DiffMix. The strongest comparator by full-budget AULC is colored and named; others are gray, and dashed lines show Random replays.}
\label{fig:curves}
\end{figure}

\method{} ranks first by mean full-budget AULC in \eqref{eq:aulc} on Drug (0.738) and ILThermo (0.748) and second on NIST (0.444) and DiffMix (0.836) (Table~\ref{tab:main}). Across paired comparisons matched by task and seed, both paired tests support higher AULC than seven of the eight comparator strategies, with \method{} winning 75 of 78 paired comparisons against those seven strategies (80 of 87 overall), median gains of $+0.031$ to $+0.265$, and Holm-adjusted $p\le0.039$ (Supplementary Table~S6). Against GP-Greedy, \method{} achieves higher AULC in five of nine pairs, though the difference is not statistically significant (Holm-adjusted $p=0.50$ for the sign test and $p=0.18$ for the Wilcoxon test).

\begin{table}[!htbp]
\centering
\caption{Full-budget AULC (primary measure): mean $\pm$ sample standard deviation (SD) over three runs, including Random replays on all four tasks. n/r: not run. Dark pink and bold mark the best value in each column; light pink marks the second best.}
\label{tab:main}
\scriptsize\setlength{\tabcolsep}{2.6pt}
\begin{tabular}{@{}lcccc}
\toprule
Strategy & \shortstack{\strut NIST\\\cite{kazakov2012source}} & \shortstack{\strut Drug\\\cite{bao2024solubility}}
& \shortstack{\strut ILThermo\\\cite{dong2007ilthermo}} & \shortstack{\strut DiffMix\\\cite{diffmix}}\\
\midrule
\method{} (ours) & \secondbest{0.444$\pm$0.007} & \best{0.738$\pm$0.014} & \best{0.748$\pm$0.024} & \secondbest{0.836$\pm$0.025} \\
MolSets-UCB~\cite{molsets} & 0.309$\pm$0.013 & 0.650$\pm$0.023 & 0.709$\pm$0.009 & 0.326$\pm$0.055 \\
MolSets-Greedy~\cite{molsets} & 0.403$\pm$0.011 & 0.659$\pm$0.011 & 0.691$\pm$0.045 & n/r \\
SolvGNN-UCB~\cite{solvgnn} & 0.380$\pm$0.011 & \secondbest{0.682$\pm$0.015} & 0.687$\pm$0.040 & 0.815$\pm$0.034 \\
SolvGNN-Greedy~\cite{solvgnn} & 0.440$\pm$0.012 & \secondbest{0.682$\pm$0.010} & 0.720$\pm$0.033 & n/r \\
GP-EI~\cite{rasmussen2006gaussian} & 0.122$\pm$0.011 & 0.665$\pm$0.034 & 0.212$\pm$0.057 & 0.625$\pm$0.035 \\
GP-Greedy~\cite{rasmussen2006gaussian} & \best{0.476$\pm$0.009} & 0.653$\pm$0.015 & \secondbest{0.734$\pm$0.066} & n/r \\
XGBoost-UCB~\cite{chen2016xgboost} & 0.399$\pm$0.008 & 0.590$\pm$0.021 & 0.687$\pm$0.039 & 0.790$\pm$0.028 \\
XGBoost-Greedy~\cite{chen2016xgboost} & 0.401$\pm$0.019 & 0.630$\pm$0.032 & 0.672$\pm$0.029 & \best{0.838$\pm$0.016} \\ \midrule
Random & 0.028$\pm$0.003 & 0.228$\pm$0.039 & 0.081$\pm$0.039 & 0.132$\pm$0.022 \\
 
\bottomrule
\end{tabular}
\end{table}

Query efficiency ($q_{30}$ and $q_{80}$) reflects the distinct chemical regimes of the four tasks (Table~\ref{tab:efficiency}). On the 86-drug Drug task, \method{} achieves the lowest $q_{30}$ (84) and reaches 80\% macro active recall in a median of 348 queries, saving 76 queries (18\%) over the fastest comparator (SolvGNN-UCB, 424) and over 130 queries relative to descriptor surrogates (GP-Greedy, 483; XGBoost-Greedy, 569). On ILThermo, greedy selection starts faster, and \method{}'s recall curve overtakes the greedy comparators after about one-third of the budget (Fig.~\ref{fig:curves}(c)), achieving the best median $q_{80}$ (112 vs.\ 114 for GP-Greedy). On DiffMix ($\mathrm{LiPF_6}$--carbonate electrolytes), \method{} shares the lowest $q_{30}$ (30) and reaches $q_{80}=98$ (vs.\ 93 for XGBoost-Greedy); on NIST, 80\% recall is reached in three GP-Greedy runs ($q_{80}=689$) and one \method{} run ($772$; Supplementary Table~S5).

\begin{table}[!htbp]
\centering
\caption{Estimated new formulation queries to reach 30\% ($q_{30}$) and 80\% ($q_{80}$) active recall. Values are medians of interpolated counts among runs reaching each level. $^{\dagger}$/$^{\ddagger}$: one/two of three runs reach it; ---: none; n/r: not run. Drug uses equal-drug macro active recall. Shading follows Table~\ref{tab:main}, ranking more runs reaching the level first, then fewer queries.}
\label{tab:efficiency}
\footnotesize
\setlength{\tabcolsep}{1.2pt}
\renewcommand{\arraystretch}{1.12}
\begin{tabular*}{\columnwidth}{@{\extracolsep{\fill}}l*{8}{c}@{}}
\toprule
& \multicolumn{2}{c}{\shortstack{\strut NIST\\\cite{kazakov2012source}}}
& \multicolumn{2}{c}{\shortstack{\strut Drug\\\cite{bao2024solubility}}}
& \multicolumn{2}{c}{\shortstack{\strut ILThermo\\\cite{dong2007ilthermo}}}
& \multicolumn{2}{c}{\shortstack{\strut DiffMix\\\cite{diffmix}}}\\
\cmidrule(lr){2-3}\cmidrule(lr){4-5}\cmidrule(lr){6-7}\cmidrule(l){8-9}
Strategy & $q_{30}$ & $q_{80}$ & $q_{30}$ & $q_{80}$
& $q_{30}$ & $q_{80}$ & $q_{30}$ & $q_{80}$\\
\midrule
\method{} (ours) & 267 & \secondbest{772$^{\dagger}$} & \best{84} & \best{348} & 39 & \best{112} & \best{30} & \secondbest{98} \\
MolSets-UCB~\cite{molsets} & 392 & --- & 113 & 446 & 39 & 143 & 229 & 375 \\
MolSets-Greedy~\cite{molsets} & 288 & --- & 105 & 486 & \best{29} & 150 & n/r & n/r \\
SolvGNN-UCB~\cite{solvgnn} & 294 & --- & \secondbest{100} & \secondbest{424} & 54 & 134 & 63 & 108 \\
SolvGNN-Greedy~\cite{solvgnn} & \secondbest{255} & --- & 103 & 436 & 38 & 128 & n/r & n/r \\
GP-EI~\cite{rasmussen2006gaussian} & --- & --- & 116 & 473 & 246$^{\dagger}$ & --- & \secondbest{34} & 287$^{\ddagger}$ \\
GP-Greedy~\cite{rasmussen2006gaussian} & \best{233} & \best{689} & 111 & 483 & \secondbest{34} & \secondbest{114} & n/r & n/r \\
XGBoost-UCB~\cite{chen2016xgboost} & 296 & --- & 125 & 575 & 51 & 145 & 38 & 129 \\
XGBoost-Greedy~\cite{chen2016xgboost} & 274 & --- & 113 & 569 & 43 & 161 & \best{30} & \best{93} \\ \midrule
Random & --- & --- & 464 & --- & --- & --- & 355$^{\dagger}$ & --- \\
 
\bottomrule
\end{tabular*}
\end{table}

\begin{figure}[!htbp]
\centering
\includegraphics[width=\columnwidth]{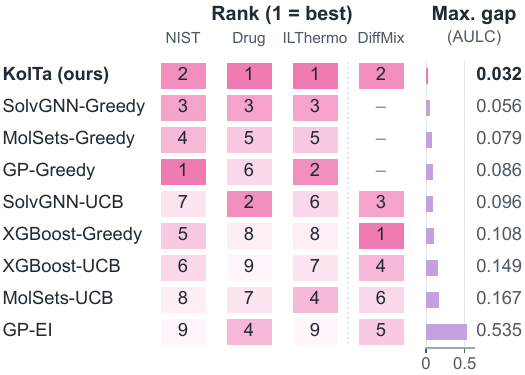}
\caption{\method{} has the smallest maximum mean-AULC gap across the three measured tasks (0.032 vs.\ $\ge$0.056 for every comparator). Left: ranks by three-run mean full-budget AULC (1: best; darker: better) among nine strategies on measured tasks and six on DiffMix (--: not run). Right: worst-case gap to the best mean AULC across the three measured tasks (\method{} is 1st or 2nd on every task).}
\label{fig:consistency}
\end{figure}

\textbf{Cross-task consistency and locked-set accuracy.}\label{sec:consistency} Across NIST, Drug, and ILThermo, \method{} places in the top three of nine strategies in all nine runs, compared with seven for SolvGNN-Greedy (Supplementary Table~S8(a)), and achieves the smallest worst-case AULC gap: 0.032, versus at least 0.056 for every comparator (Fig.~\ref{fig:consistency}). Comparator performance fluctuates sharply across regimes: GP-Greedy ranks first on NIST but sixth on Drug; SolvGNN-UCB ranks second on Drug but seventh on NIST; GP-EI drops to 0.122 on NIST and 0.212 on ILThermo; and on DiffMix, RMS-ratio MolSets-UCB drops to 0.326 as its exploration bonus redirects early batches away from the highest predicted means (zero top-40 overlap in eight of the twelve batches pooled from the first four rounds of the three runs; Supplementary Section~S3.1). On the locked sets, \method{} ranks in the top three in final MAE across all four tasks (first on Drug at 0.173, second on ILThermo at 0.134, and third on NIST at 0.108 and DiffMix at 0.051), whereas GP-EI attains the lowest global MAE on three tasks while lagging in top-5\% target discovery (Supplementary Table~S10).

\subsection{Predictor Comparison}\label{sec:sources}
To isolate the predictor from acquisition calibration, we compare all model families under a common greedy rule ($s\,\mu(F)$) on Drug and ILThermo (Table~\ref{tab:ablation}(b)). The \method{} predictor (A2) achieves both the highest mean AULC and lowest median $q_{80}$ on each task: 0.712 ($q_{80}=383$) on Drug vs.\ 0.682 ($436$) for SolvGNN-Greedy, and 0.763 ($q_{80}=108$) on ILThermo vs.\ 0.734 ($114$) for GP-Greedy. Across the six paired runs, A2 achieves higher AULC than SolvGNN-Greedy, MolSets-Greedy, and XGBoost-Greedy in all six and GP-Greedy in five (Supplementary Tables~S4 and~S7), showing that the three-stage physics-informed predictor provides a stronger surrogate even before adding exploration.

\begin{table}[!htbp]
\centering
\caption{Acquisition and predictor comparisons: full-budget AULC (mean $\pm$ sample SD over three runs) and median estimated $q_{80}$. (a) Acquisition variants sharing the \method{} predictor and physics training ($c_v=0$ in A1). (b) Greedy selection ($s\,\mu$ only) across predictors, starting with A2. Shading follows Table~\ref{tab:main}.}
\label{tab:ablation}
\scriptsize\setlength{\tabcolsep}{2.4pt}
\begin{tabular}{@{}lcccc}
\toprule
 & \multicolumn{2}{c}{\shortstack{\strut Drug\\\cite{bao2024solubility}}} & \multicolumn{2}{c}{\shortstack{\strut ILThermo\\\cite{dong2007ilthermo}}}\\
\cmidrule(lr){2-3}\cmidrule(l){4-5}
 & AULC & $q_{80}$ & AULC & $q_{80}$\\
\midrule
\multicolumn{5}{@{}l}{\emph{(a) Acquisition variants on the \method{} predictor}} \\
\quad \method{} (ours; PRC-UCB) & \best{0.738$\pm$0.014} & \best{348} & \secondbest{0.748$\pm$0.024} & \secondbest{112} \\
\quad A1: margin with $u$ only ($c_v=0$) & \secondbest{0.729$\pm$0.005} & \secondbest{360} & 0.736$\pm$0.010 & 115 \\ \midrule
\multicolumn{5}{@{}l}{\emph{(b) Predictor comparison under greedy selection ($s\,\mu$ only)}} \\
\quad \method{} (ours) predictor (A2) & 0.712$\pm$0.018 & 383 & \best{0.763$\pm$0.039} & \best{108} \\
\quad SolvGNN-Greedy~\cite{solvgnn} & 0.682$\pm$0.010 & 436 & 0.720$\pm$0.033 & 128 \\
\quad MolSets-Greedy~\cite{molsets} & 0.659$\pm$0.011 & 486 & 0.691$\pm$0.045 & 150 \\
\quad GP-Greedy~\cite{rasmussen2006gaussian} & 0.653$\pm$0.015 & 483 & 0.734$\pm$0.066 & 114 \\
\quad XGBoost-Greedy~\cite{chen2016xgboost} & 0.630$\pm$0.032 & 569 & 0.672$\pm$0.029 & 161 \\
 
\bottomrule
\end{tabular}
\end{table}

\subsection{Acquisition Analysis}\label{sec:acquisition_results}
Comparing greedy A2, disagreement-only A1 ($c_v=0$), and full PRC-UCB (\method{}) in \eqref{eq:acq} shows how calibrated exploration and the physics residual improve discovery and cross-task consistency (Table~\ref{tab:ablation}(a); Supplementary Table~S7). Adding the physics residual $v(F)$ increases mean AULC over $u$-only A1 on both tasks, rising from 0.729 to 0.738 on Drug ($q_{80}$: 360 to 348) and from 0.736 to 0.748 on ILThermo ($q_{80}$: 115 to 112), and winning five of six paired runs (all three ILThermo seeds by $+0.003$ to $+0.027$ and two of three Drug seeds by $+0.018$ and $+0.022$). Across Drug and ILThermo, PRC-UCB also attains the smallest worst-case AULC gap among the three variants (0.015, vs.\ 0.026 for A2 and 0.027 for A1; Supplementary Table~S8(b)) and lowers AULC standard deviation relative to A2 on both tasks.

The OOF margin coefficients fitted via \eqref{eq:pinball}--\eqref{eq:unscale} across all 120 acquisition rounds (30 per task; Supplementary Table~S9) show that both signals remain active across tasks. Across the 120 rounds, $c_u>0$ in all 120 and $c_v>0$ in 115 (29/30 on NIST and Drug, 30/30 on ILThermo, and 27/30 on DiffMix). In RMS-scaled coordinates $(c'_u,c'_v)=(c_u\sigma_u,c_v\sigma_v)$, the median fitted weights are $(0.235,0.038)$ on Drug (largest $c'_u$), $(0.122,0.052)$ on ILThermo (largest $c'_v$), $(0.071,0.025)$ on NIST, and $(0.037,0.008)$ on DiffMix. On Drug, where all variants follow the same fixed, balanced schedule to select formulations within each scheduled drug, adding the calibrated margin lifts mean AULC from 0.712 for greedy A2 to 0.729 for A1 and 0.738 for PRC-UCB (winning all three seeds over A2).

\section{Discussion and Limitations}\label{sec:discussion}

\textbf{Adapting the framework.} \method{} separates its learning and calibration procedure from task-specific component roles and physical relations. Roles determine which components enter the base readout and which directed interactions are modeled; the anchored temperature network and derivative residual guide the learned temperature response without restricting reference-temperature predictions to a closed-form curve. Sections~\ref{sec:discovery} and~\ref{sec:sources} show that this modular design outperforms general graph and descriptor baselines under common greedy selection and maintains top-two discovery performance across all four tasks while comparator rankings fluctuate across chemical domains.

\textbf{Role of physics and OOF calibration in acquisition.} Our results show that low global prediction error alone does not ensure fast target discovery: GP-EI attains the lowest final locked-set MAE on NIST, ILThermo, and DiffMix, while ranking ninth, ninth, and fifth, respectively, in mean discovery AULC, and on DiffMix the RMS-ratio exploration bonus in MolSets-UCB redirects early batches away from top-mean candidates (Supplementary Section~S3.1). Conversely, under the fixed, balanced drug schedule on Drug, pure greedy selection within each scheduled drug group (A2, 0.712) trails calibrated exploration (A1, 0.729; PRC-UCB, 0.738). Rather than using a single RMS-ratio weight, PRC-UCB fits nonnegative weights for ensemble disagreement $u(F)$ and the target-temperature physics residual $v(F)$ jointly to OOF signed utility errors via asymmetric pinball loss ($\tau=0.8$), assigning positive weight to $v(F)$ in 115 of 120 rounds (Supplementary Table~S9) and improving AULC over $u$-only A1 in five of six runs.

\textbf{Limitations.} First, the current physics loss applies soft constraints at five fixed temperatures per task, and extrapolation beyond the training temperature range has not been evaluated. Incorporating physical relations directly into the temperature-response function is a possible direction for improving extrapolation. Second, the model treats temperature as its only varying condition; extending it to jointly varying temperature and pressure would require additional condition inputs and coupled physical constraints.

\section{Conclusion and Future Work}\label{sec:conclusion}
We have presented \method{}, a physics-informed active learning framework for discovering high-performing formulations from limited measurements. It combines a three-stage predictor trained on all measured temperature records with task-specific soft physics constraints and PRC-UCB acquisition. Across four benchmarks spanning binary mixtures, drug solutions, ionic liquids, and electrolytes, \method{} consistently ranks in the top two strategies by discovery AULC without retuning its acquisition rule across domains. Under a common greedy rule, its three-stage physics-informed predictor outperforms graph and descriptor surrogates on both ablation tasks, while OOF calibration of ensemble disagreement and the target-temperature physics residual improves mean AULC over disagreement-only calibration on both tasks and achieves the smallest worst-case AULC gap across the acquisition variants (0.015 vs.\ 0.026 for greedy A2 and 0.027 for disagreement-only A1). Code will be made publicly available upon acceptance.

Future work will adapt the framework to more material systems and test it in laboratory experiments, investigate incorporating physical relations directly into the temperature-response function for temperature extrapolation, extend the model to jointly varying temperature and pressure, and explore richer multimodal molecular representations.

\section*{Conflict of Interest}
The authors declare that they have no conflicts of interest.
\label{bodyend}
\clearpage
\bibliographystyle{IEEEtran}
\bibliography{references}
\end{document}